\pdfoutput=1

\documentclass[11pt]{article}

\usepackage{color,soul}
\usepackage{EMNLP2023}
\usepackage{fontawesome5}

\usepackage{multirow}
\usepackage{booktabs}
\usepackage{subfig}
\usepackage{float}
\usepackage{bbm}
\usepackage{float}
\usepackage{arabtex}
\usepackage{utf8}
\setcode{utf8}
\usepackage{times}
\usepackage{amsmath}
\usepackage{latexsym}
\usepackage{lipsum}
\usepackage{color}
\usepackage{xcolor}
\usepackage{colortbl}
\usepackage{graphicx}

\usepackage[T1]{fontenc}

\usepackage[utf8]{inputenc}

\usepackage{microtype}

\usepackage{inconsolata}

\title{Violet: A Vision-Language Model for Arabic Image Captioning with Gemini Decoder}

\author{\normalsize  Abdelrahman Mohamed$^{\xi}$~ Fakhraddin Alwajih$^{\lambda}$~El Moatez Billah Nagoudi$^{\lambda}$ \\
\normalsize\textbf{Alcides Alcoba Inciarte}$^{\lambda}$ ~ \normalsize\textbf{Muhammad Abdul-Mageed}$^{\lambda,\xi}$~  
 \\
\normalsize $^{\lambda}$ Deep Learning \& Natural Language Processing Group,
  The University of British Columbia\\\normalsize  $^{\xi}$Department of Natural Language Processing \& Department of Machine Learning, MBZUAI\\ %
  \texttt{\normalsize \{fakhr.alwajih,moatez.nagoudi,muhammad.mageed\}@ubc.ca}}

\begin{document}
\maketitle

\begin{abstract}

Although image captioning has a vast array of applications, it has not reached its full potential in languages other than English. Arabic, for instance, although the native language of more than 400 million people, remains largely underrepresented in this area. This is due to the lack of labeled data and powerful Arabic generative models. We alleviate this issue by presenting a novel vision-language model dedicated to Arabic, dubbed \textit{Violet}. Our model is based on a vision encoder and a Gemini text decoder that maintains generation fluency while allowing fusion between the vision and language components. To train our model, we introduce a new method for automatically acquiring data from available English datasets. We also manually prepare a new dataset for evaluation. \textit{Violet} performs sizeably better than our baselines on all of our evaluation datasets. For example, it reaches a CIDEr score of $61.2$ on our manually annotated dataset and achieves an improvement of $13$ points on Flickr8k.

\end{abstract}

\section{Introduction}\label{intro}
\begin{figure}[htp!]
  \centering
    \subfloat[\centering\<امرأة وفتاة يلعبان  بالفريزبي\\ على العشب>]{\label{figur:3}\includegraphics[width=35mm, height=30mm]{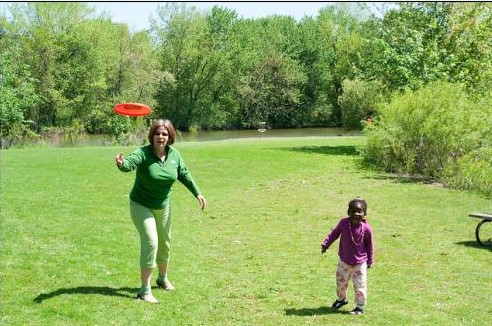}}
 \hspace{2mm}
  \subfloat[\centering\<كلب أبيض وأسود يسبح\\ في الماء>]{\label{figur:2}\includegraphics[width=35mm, height=30mm]{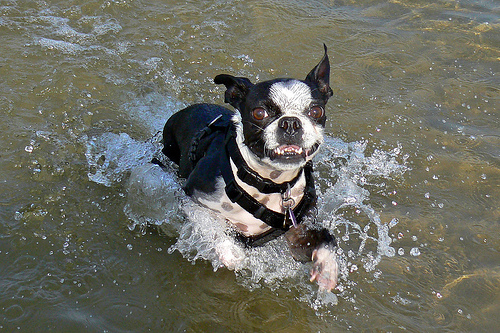}}\\[-1ex]
   \subfloat[\centering\<قطة مستلقية بجانب
  جهاز \\تحكم عن بعد> ]{\label{figur:1}\includegraphics[width=35mm, height=30mm]{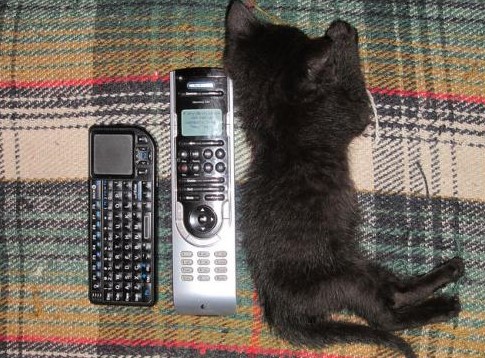}} 
   \hspace{2mm} 
  \subfloat[\centering\<كعكة عيد ميلاد مع هاتف\\ محمول عليها>]{\label{figur:4}\includegraphics[width=35mm, height=30mm]{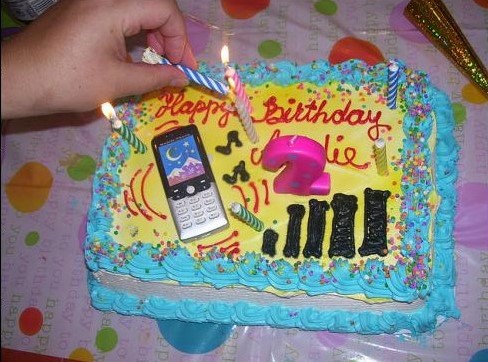}}\hspace{2mm}
\caption{Examples of captions generated by our model.}
 \label{figur}

\end{figure}

Captioning images involves describing the visual elements of a picture using natural language. This requires a system that combines the strengths of two models: one that can represent the visual elements of an image, and another that can translate this representation into natural language. The latter employs a language model to produce \textit{fluent} (i.e., grammatically accurate) and \textit{adequate} (i.e., capturing sufficient semantic information) descriptions. In recent years, research on vision language models (VLMs) and their applications has boomed \cite{alayrac2022flamingo,wang2022git,huang2023language}. Owing to the rapid advancements in large language models (LLMs), the performance of VLMs has improved dramatically. More concretely, VLMs have progressed from merely providing descriptions that vaguely resemble a given image~\cite{vinyals2015show} to accurately describing complex visual cues within the image. The \textit{pretraining-then-finetuning} paradigm also plays a significant role in achieving such impressive results, as it allows models to first grasp general language structures and then specialize in the specific task of image captioning~\cite{gan2022vision}.

Progress in VLMs, however, has been witnessed thus far primarily on English~\citet{awais2023foundational}. This leaves behind a large number of other languages for which no sufficient image captioning data or language models exist. Arabic is a case in point where image captioning lags far behind \cite{elbedwehy2023improved}. 
Similar to other low-resource languages, progress in Arabic image captioning has been hampered by the lack of publicly available datasets and limited efforts in creating any such data. Manual creation of image datasets, after all, requires a huge amount of time and labor. Again, the unavailability of powerful Arabic language models that understands the structure of the language and can capture its rich morphology has also caused a delay in the development of VLMs. Given the rapid progress in vision language technologies and their wide applications in society, limited progress in this area can have negative consequences for the Arabic-speaking world. 

\begin{figure}[htp]
    \centering
    \includegraphics[width=0.48
\textwidth]{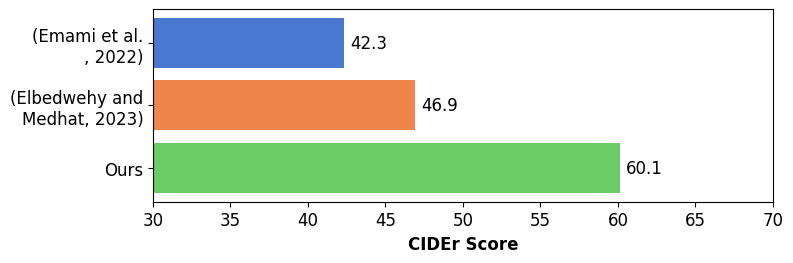}
    \caption{Performance of our model compared to previous works on Flickr8k using CIDEr metric.}
    \label{fig:comp}

\end{figure}

To address this important issue, we introduce a novel Arabic image captioning model dubbed \textit{Violet}. Our new model is comprised of two main components: a vision encoder and a text decoder. For the vision encoder, we employ an object detector network based on FasterRCNN \cite{ren2015faster} to extract visual features that are then passed to a compact transformer encoder. At the decoder side, we leverage the recently developed generative pretrained model JASMINE \cite{nagoudi2022jasmine}. Taking inspiration from~\cite{yu2022coca}, we split our text decoder into two halves: the first half functions as a text decoder, whereas the second incorporates cross-attention layers, effectively serving as a fusion decoder. Given the dual nature of our decoder, we refer to it as \textit{Gemini}. Drawing parallels with VisualGPT~\cite{chen2022visualgpt} and the meshed transformer~\cite{cornia2020meshed}, we also adopt a meshed connection between the transformer vision encoder and the text decoder to foster enhanced communication between the encoder and decoder layers.
\begin{figure*}[htp]
    \centering
    \includegraphics[width=0.92
\textwidth]{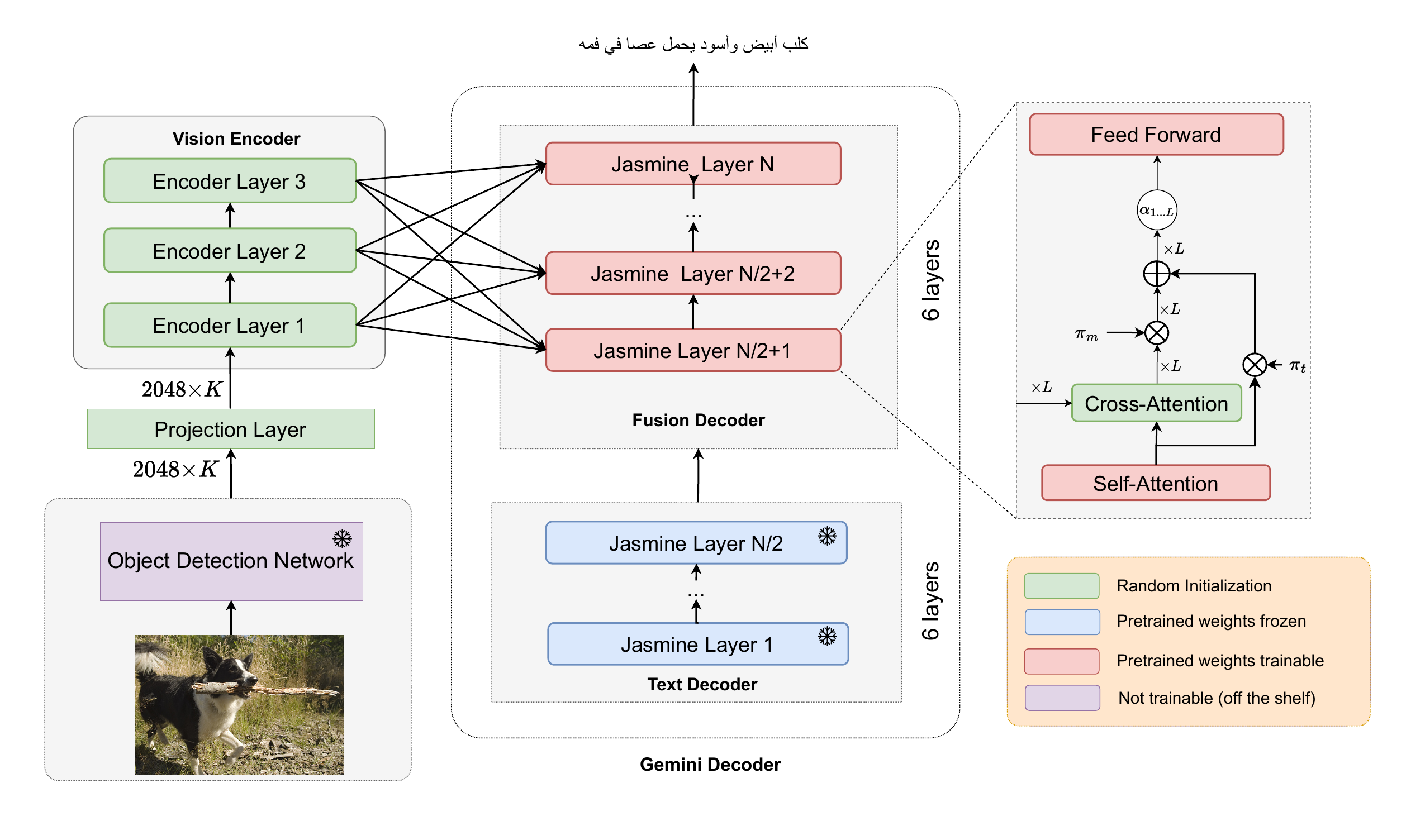}
    \caption{The architecture and output generated by our model. We use an object detection network to extract K object features (K equal $50$ in our case)  from an image. After projecting to a lower dimension, the features are fed to an L-layer (three-layer in our architecture) transformer encoder. Meshed connection is employed between the encoder and decoder layers, where each encoder layer contributes to the cross-attention output. Our text decoder is split into two halves, the first half is the standard frozen pretrained text decoder layers, while the second half has cross-attention layers inserted after each self-attention layer. We call this design a \textit{Gemini decoder}. We employ a gating mechanism through $\pi_t$ and $\pi_m$ that controls the flow of information from the vision and language sides.  The final input to the feed forward network in each cross-attention layer is the weighted sum of each encoder-decoder attention controlled by the $\alpha$ parameters.}
    \label{fig:arch}
\end{figure*}

The other major challenge we face in our work is the unavailability of native Arabic captioning data. We alleviate this challenge by introducing a method for automatically acquiring captions that is based on first employing a powerful machine translation model followed by a quality assurance mechanism for removing poor captions. For evaluation, in addition to reporting on Arabic translated dataset, we task five human annotators to manually caption an image dataset. Compared to previous works and baselines, our novel model excels in captioning images in fluent Arabic. Figure \ref{figur} offers four examples of fluent Arabic captions generated by our novel model. Figure \ref{fig:comp} shows a comparison of our model performance with prior research on Flickr8k in CIDEr score.   

\noindent In summary, our contributions are as follows:
\begin{itemize}\setlength{\itemsep}{0em}
    \item We present a novel image captioning model that employs an effective pretrained Arabic decoder capable of outputting rich captions.
    \item Our model achieves competitive performance for Arabic image captioning on both the MSCOCO  \cite{lin2014microsoft} and Flicker8k \cite{flicker8k_caffe} datasets, establishing a new state-of-the-art in this area.
    \item In the process of developing our new model, we release a translated version of MSCOCO dataset that has gone through our quality assurance pipeline. Our released dataset can help further advance research in Arabic VLMs. 
    \item We also release our manually captioned dataset, a subset of MSCOCO test set, that we dub \textit{AraCOCO}. 
\end{itemize}

\section{Related Work}
\textbf{Image captioning.} Early methods for image captioning involve either retrieving descriptions~\cite{karpathy2014deep} or using template filling combined with manually designed natural language generation techniques \cite{yang2011corpus,li2011composing}. However, modern image captioning primarily relies on deep learning models. In early work, image captioning is framed as an image-to-sequence task using encoder-decoder models, with Convolutional Neural Networks (CNNs) as encoders and Recurrent Neural Networks (RNNs) as decoders while incorporating attention mechanisms \cite{xu2015show,you2016image,huang2019attention}. Soon after, using a transformer architecture of a vision encoder with a text decoder became the defacto direction towards solving the problem of image captioning \cite{stefanini2022show}. Some approaches use a detection model to extract visual features and then pass it to a transformer text decoder as in Oscar \cite{li2020oscar,chen2022visualgpt}, while others like CoCa~\cite{yu2022coca} train a transformer vision encoder with a text decoder from scratch on a large-scale dataset.

More recently, there has been a shift towards using pre-trained LLMs and vision models. Generative Image-to-text Transformer (GIT) \cite{wang2022git} is a decoder-only transformer that utilizes a CLIP~\cite{radford2021learning} visual encoder to incorporate both visual and textual inputs. Another method to consider is VisualGPT~\cite{chen2022visualgpt} which uses a pretrained FasterCNN to extract visual features that it passes to a small vision encoder. For the decoder side, it uses the text-pretrained model GPT2 \cite{radford2019language}. 

\noindent\textbf{Arabic image captioning.} Arabic poses significant challenges to image captioning. This is due to the lack of native Arabic captioning datasets in the public domain, the morphological complexity of Arabic, and the large number of diverse dialects \cite{attai2020survey}. However, a number of Arabic image captioning works exist. For instance, approaches such as root-word based RNNs and deep neural networks are used for direct Arabic caption generation \cite{jindal2017deep}.~\citet{al2018automatic} employ a generative merge model with three components: an LSTM-based language model, a CNN-based image feature extraction model, and a decoder that processes outputs from the first two models. \citet{eljundi2020resources} introduce an Arabic captioning model trained on a translated Flickr8K dataset, discussing issues related to translation.~\citet{afyouni2021aracap} present AraCap, a hybrid design that combines a CNN with object detection using attention mechanisms and produces captions through an LSTM. They train their model on MSCOCO and Flickr30k \cite{plummer2015flickr30k} datasets and test on an Arabic translated subset of MSCOCO.~\citet{lasheen2022arabic} propose an encoder-decoder structure, incorporating attention mechanisms with CNN encoding and LSTM decoding. In another study~\cite{emami2022arabic}, various Arabic image captioning models are formulated and assessed using standard metrics. The authors use transformers pretrained on diverse Arabic datasets following the architecture and training method introduced in OSCAR ~\cite{li2020oscar}.
~\newcite{elbedwehy2023improved} present a model employing transformers for both encoding and decoding. It uses feature extraction from images in the encoding stage and a pretrained word embedding model in the decoding stage, all tested on the Arabic-translated Flickr8k dataset in~\citet{eljundi2020resources}. This work is closest to ours in that we also utilize transformer encoders and decoders. However, we use a GPT-styled decoder that endows our approach with high Arabic fluency.


\section{Approach}
\subsection{Model Architecture}
Our model is a vision-encoder-decoder architecture. For the vision encoder part, we employ an object detection network \cite{anderson2018bottom} and a three-layer transformer. For the text decoder, we use the pretrained transformer decoder JASMINE \cite{nagoudi2022jasmine}. To align visual and textual features, we utilize cross-attention. In standard attention, also known as self-attention, the attention output is computed using three matrices derived from the same input: the query matrix $Q$, the key matrix $K$, and the value matrix $V$. More concretely, given an input sequence represented as a matrix \( S_t \), where each row corresponds to a vector in the sequence, the attention is
calculated as:
\begin{equation}
\small
\begin{aligned}
Attn(S_t) = \text{softmax}\left(\frac{S_t W_q (S_t W_k)^T}{\sqrt{d_k}}\right) S_t W_v
\end{aligned}
\end{equation}

Where \( W_q \), \( W_k \), and \( W_v \) are the learnable weight matrices for the query \( Q \), key \( K \), and value \( V \) respectively. \( d_k \) is the dimensionality of the query/key vectors. The division by \( \sqrt{d_k} \) is a scaling factor to ensure the dot products don't grow too large as the dimensionality increases.

In the case of cross-attention, the query is derived from the output of the text decoder's self-attention, while the key and value are sourced from the vision encoder. Mathematically, given image visual features output $S_m$, and the textual features $S_t$, the formula becomes:

\begin{equation}
\small
\begin{aligned}
XAttn(S_t,S_m) &= \text{softmax}\left(\frac{(S_tW_q)(S_mW_k)^T}{\sqrt{d_k}}\right) \\
&\times S_mW_v
\end{aligned}
\end{equation}


\noindent Now that the attention mechanism foundations are laid out, we describe our vision encoder and text decoder in detail.
 
\subsubsection{Vision Encoder}
Our vision encoder consists of two components: a pretrained object detection network, and a three-layer transformer encoder. For the object detection network, we employ bottom-up attention network \cite{anderson2018bottom}. In our initial experiments, it results in superior visual features compared to using the vanilla FasterRCNN model \cite{ren2015faster}. Previous works \cite{li2020oscar,cornia2020meshed,chen2022visualgpt} also show the effectiveness of this network in feature extraction.
The transformer encoder, on the other hand, is a three-layer standard transformer architecture that takes the output of the detection network to further refine the visual features. For each image, the detection network detects the potential objects and extracts the visual features from their bounding boxes.\footnote{A bounding box is a region in the image that contains the object.} These visual features are passed through a projection layer and then fed to the three-layer transformer encoder as input. We adapt meshed connection~\cite{cornia2020meshed} in our architecture between the encoder layers and the text decoder. This allows all the encoder layers to contribute to the input of the cross-attention rather than using only the output of the last encoder layer. The contribution of each encoder layer is determined by the learnable parameters matrix $\alpha$. For each layer $i$, $\alpha_i$ is calculated as:


\begin{equation}
\small
\begin{aligned}
\alpha_i = \sigma(W_i[S_t\parallel XAttn(S_{m_i},S_t)+b_i])
\end{aligned}
\end{equation}

Where $S_t$ is the input sequence of each decoder layer, $\sigma$ is the sigmoid activation function, $W_i$ is a learnable weight matrix, $b_i$ is a bias term and $\parallel$ indicates concatenation. This measures the relevance between the input for each decoder layer $S_t$, and the output of each encoder layer.


\subsubsection{Gemini Decoder}
We employ the pretrained Arabic decoder JASMINE \cite{nagoudi2022jasmine} as our text decoder. JASMINE  is a decoder-based transformer that follows  GPTNeo architecture  \cite{black_sid_2021_5551208}. JASMINE models range in complexity from $300$ million to $13$ billion parameters and are trained on a text dataset of approximately $400$GB, covering diverse Arabic varieties from multiple domains. We utilize the JASMINE base variant in our architecture, which is a 12-layer transformer decoder with a $768$-dimensional embedding.


Although the meshed connection introduced in~\citet{cornia2020meshed} proved to have positive improvements on performance due to the richer visual features, calculating the cross-attention of each encoder layer with each decoder layer is computationally expensive.
Inspired by~\citet{yu2022coca}, we split our pretrained text decoder into two parts. The first part acts as a vanilla text decoder, while the second part acts as a fusion decoder that aligns visual and textual features.
This design choice serves two purposes. First, it reduces the computations and the number of parameters by removing cross-attention layers and the mesh connections in the first half of the decoder. Second, having its first half intact acting as a vanilla text decoder, allows our decoder to keep its innate generative capabilities, while also enabling smoother convergence. 

As shown in Figure \ref{fig:arch}, the first half has only the pretrained self-attention layers of JASMINE. While the second half got cross-attention blocks inserted in-between each layer, acting as a fusion decoder. To ensure maintaining the functionality of our pretrained decoder, we freeze the first part that acts as the text decoder. This modification not only decreases computational cost but also positively impacts overall performance.
In order to further enhance the quality of the features generated by both the vision encoders and the text decoder, we employ self-resurrecting activation unit (SRAU) introduced in~\citet{chen2022visualgpt}.  The process of generating a caption relies on visual cues to convey the image's content and textual cues to provide relationships between words for a coherent and fluent output. To allow the important information to flow without distortion, SRAU selectively permits the activation above a certain threshold through a gating mechanism. This effectively filters out any weak signal produced by either the vision or language part.

Concretely, as shown in Figure \ref{fig:arch}, for each encoder-decoder connection, the output $Z_i$ to the feedforward layer is calculated as:

\begin{equation}
\small
    Z_i = \pi_m\otimes XAttn(S_t,S_{m_i})+ \pi_t\otimes Attn(S_t),
\end{equation}
in which  $\pi_m$ is the gating parameter for the vision part and $\pi_t$ for the text part, calculated as:
{\small

\begin{align*}
\pi_m &=\sigma(A_n) \mathbbm{1} (\sigma(A_n) > \tau), &\forall n \in Attn(S_t) \\
\pi_t &= (1-\sigma(A_n)) \mathbbm{1} (1-\sigma(A_n) >\tau) &\forall n \in Attn(S_t)
\end{align*}
}%

\noindent where $\sigma$ is the sigmoid function, $A_n$ is an element in the attention matrix, $\mathbbm{1}$ is an indicator function that equals one if the condition is true and zero otherwise, and $\tau$ is a hyperparameter. This negates any disturbance caused by weak activations below the threshold $\tau$ by zeroing them out.
The final output $Z$ to the feedforward layer will be the sum of each encoder-decoder connection weighted by the learned parameter $\alpha$ introduced earlier, mathematically:

\begin{equation}
\small
    Z = \frac{1}{\sqrt{L}} \sum_{i=1}^L \alpha_i Z_i
\end{equation}
Where $L$ is the number of encoder layers, set to three in our architecture.



\begin{figure}[!htp]
    \centering
    \includegraphics[width=0.48\textwidth, height=0.28\textheight]{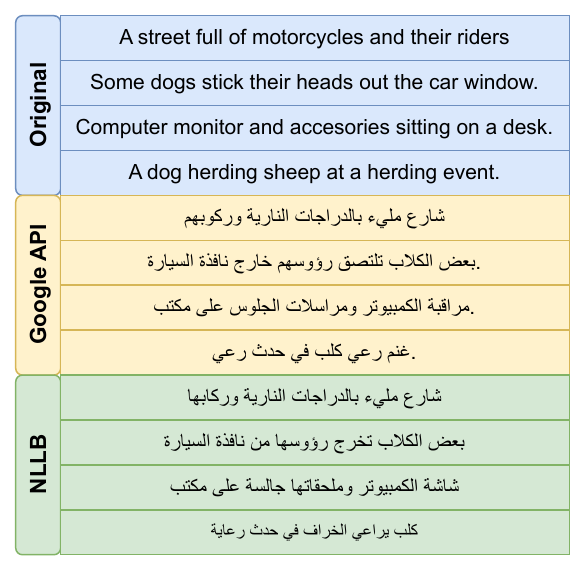}
    \caption{A comparison between the translations produced by Google translate API and NLLB for MSCOCO dataset. Unlike NLLB, Google API tends to give literal translations without incorporating the context.}
    \label{fig:com2}
\end{figure}

\subsection{Data Collection}

Owing to the unavailability of high-quality Arabic captioning training data, we first start by creating a training dataset for our model. Manually labeling and creating a new dataset would be both time-consuming and expensive; therefore, we opt for translating the commonly used captioning dataset Microsoft Common Objects in Context (MSCOCO) \cite{lin2014microsoft}. There are two famous training/validation splits for this dataset, the 2014 Karpathy's split, and the 2017 split. Both splits contain the same images and only differ in the split ratio. The dataset covers around $80$ different objects in a total of $123$k images with $5$ captions per image. The dataset is annotated manually, which makes it suitable for evaluation. We create our dataset in two steps, (i) translating the English MSCOCO, followed by (ii) a quality assurance step to filter poor translations.


\subsubsection{Machine Translation} 

In all of the previous attempts at Arabic image captioning pretraining \cite{eljundi2020resources,sabri2021arabic,emami2022arabic}, Google translate API \cite{googletranslateapi2023} was used for translating the datasets. However, the quality of the translations produced by it is not satisfactory. In~\citet{sabri2021arabic} it is reported that from a random sample of $150$ examples, a whooping $46$\% of the translations obtained by Google API are unintelligible.
Motivated by that, we investigate Meta's \textit{No Language Left Behind model (NLLB)} model~\cite{costa2022no} for translation. Figure \ref{fig:com2} illustrates a comparison between the translations produced by the Google Translate API and NLLB for four sentences sampled from MSCOCO dataset.

We conduct our comparison between the two translation models, Google Translate API{\footnote{The Google Translate API, integrated into Google Sheets, was used to translate the subset of data utilized in the comparison.} and NLLB, on two aspects. First, we manually check the quality of $200$ sentences translated by both models. Second, we calculate the perplexity of the translations of both models using our JASMINE decoder. Perplexity calculates the probability of a given sequence, providing insight into the fluency of the output translations. Lower perplexity scores indicate better fluency, while higher scores indicate poor fluency. This metric helps us to quantitatively gauge how good the translations are, supplementing our manual evaluation to offer a comprehensive understanding of the models' performance. Subsequently, our observations reveal that the Google API tends to provide a more literal translation in comparison to NLLB. Empirically speaking, we find that $42$\% of Google's translations are unintelligible, a stark contrast to the mere $15$\% from NLLB. Interestingly, this observation is consistent with findings presented in~\citet{sabri2021arabic}. Furthermore, when pitted against ChatGPT \cite{ouyang2022training}, the latter displays an impressive error rate of only $7$\% in its translations. However, we opted for NLLB due to its open-source nature.
\begin{table*}[htp]

\centering
\small 
\resizebox{2\columnwidth}{37mm}{
\begin{tabular}{p{5.0cm}p{5.0cm}p{5.0cm}}
\toprule
 English Caption & NLLB Translation & AraCOCO \\
\midrule
An airport with large jetliners and a bus traveling on a tarmac. & \<مطار مع طائرات كبيرة وحافلة\\ تسافر على المدرج> & \<أشجار النخيل أمام مطار به طائرتان كبيرتان\\ للركاب وحافلات مكوكية.> \\
  a group of buses driving around at the airport & \<مجموعة من الحافلات تسير في المطار> & \<مجموعة من الحافلات تتجول\\ في المطار> \\
 Airplanes sit at the gate as transportation vehicles move about. & \<الجيران تجلس عند البوابة بينما\\ تتحرك مركبات النقل.> & \<طائرات متوقفة عند بوابة \\المطار وهناك مركبات نقل> \\
  A busy runway with buses and luggage carts driving around & \<مدرج مزدحم مع الحافلات وعربات\\ الأمتعة التي تقود حولها> & \<مدرج مزدحم مع حافلات وعربات\\ أمتعة تتجول> \\
 An airplane and busses are lined up at the airport. & \<طائرة وحافلات منتظمة \\في المطار> & \<طائرة وحافلات منتظمة\\ في المطار
> \\
\bottomrule
\end{tabular}}

\caption{A comparison between original MSCOCO captions (first column), their NLLB translations (second column), and AraCOCO captions (third column) for the image in Figure~\ref{fig:aracoco}. }
\label{tab:aracoco}
\end{table*}
\begin{figure}[!htp]
    \centering
    \includegraphics[width=0.48\textwidth]{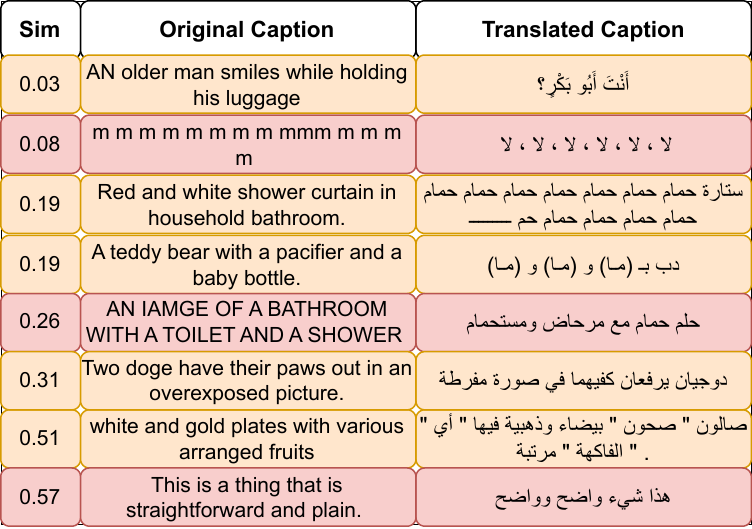}
    \caption{Examples of the rejected translations from the dataset and their semantic similarity to the English caption. Where orange highlighting refers to poor translation, and red highlighting refers to poor original caption.}
    \label{fig:comparision}

\end{figure}
\subsubsection{Data Quality Assurance} 
Although NLLB in general provides better translations compared to that of Google API, it can still output `hallucinations' and ultimately poor translations. This can be seen in the orange highlighted instances in Figure \ref{fig:comparision}. Moreover, our manual inspection reveals that some English captions in the original dataset are indeed incorrect. The MSCOCO training set can have incomprehensible samples, typos, and even unrelated captions. Examples highlighted in red in Figure \ref{fig:comparision} illustrate these poor cases. To mitigate this issue, we employ a simple method based on semantic similarity that allows us to identify and reject any such examples.

The \textit{semantic similarity} of two sentences, as the term suggests, is an indicator of the extent to which these two sentences align. A simple comparison between the embeddings of the two sentences can be obtained by passing each of them through a model and a metric such as cosine similarity can be calculated to determine how alike the two embeddings are. The smaller the angle between the two vectors, the higher the similarity score, indicating that the sentences are closer in meaning. When the sentences are in different languages, it is crucial to employ a multilingual model to generate accurate embeddings, ensuring the semantic comparison remains valid across languages.
In our experiments, we employ sentence-BERT~\cite{reimers2019sentence} to calculate the semantic similarity between each original caption and its translation. We empirically chose a similarity score threshold of $0.6$,  rejecting all captions below that threshold. This results in removing a total of $60$K samples from the whole dataset, which amounts to approximately $10$\% of the data.

\subsubsection{AraCOCO Evaluation Dataset}
\label{AraCoco}

Evaluating the performance of an Arabic captioning model presents a significant challenge due to the limited availability of human captioned data. To tackle this issue, we manually annotate a subset of $500$ images from the MSCOCO test set, dubbing our resulting dataset \textit{AraCOCO}. For each of the $500$ images, we acquire five distinct captions.
\begin{figure}[htp]
    \centering
    \includegraphics[width=0.48
\textwidth,height=0.15
\textheight]{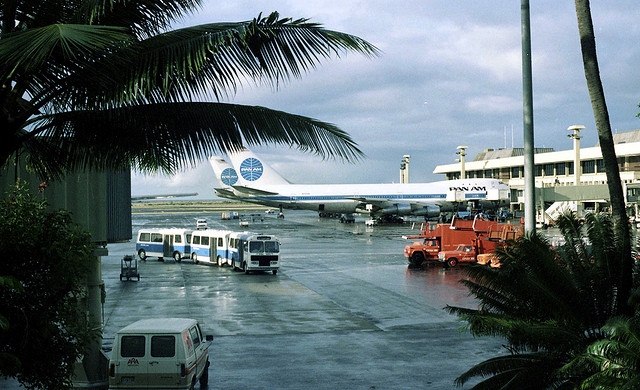}
    \caption{A sample from MSCOCO included in our AraCOCO.}
    \label{fig:aracoco}
\end{figure}
To ensure diversity of image descriptions, we acquire captions from five native Arabic-speaking annotators. The human labeling process is carried out using \textit{Label Studio}, a platform designed for such tasks.
Each annotator is presented with the same set of images and is asked to write an Arabic caption describing the image given a unique English caption as a reference. We encourage annotators to provide an Arabic caption that is more descriptive whenever possible. That is, in cases where the English caption is not capturing all details in the image, annotators are encouraged to capture these lacking details in their Arabic captions. Each annotator gets to provide only one caption per image, This approach ensures having multiple perspectives to the captions on the same image. We provide an example image from AraCOCO in Figure~\ref{fig:aracoco}, along with five different captions each acquired from one annotator in Table~\ref{tab:aracoco}. 


\begin{table*}[htp]
\centering
 \renewcommand{\arraystretch}{1.2}
\begin{tabular}{p{4.0cm}p{1.7cm}p{1.7cm}p{1.5cm}p{1.5cm}}
\toprule
\cmidrule{2-5} 
\textbf{Model}&\textbf{BLEU-1} $\uparrow$ & \textbf{BLEU-4} $\uparrow$  & \textbf{Rouge} $\uparrow$ &  \textbf{CIDEr} $\uparrow$\\
\midrule

\textbf{VisualGPT} & $56.2$ & $21.4$ & $44.1$ & $82.1$ \\ 
 \textbf{Violet (w/o Gemini)} & $45.1$& $11.3$ & $34.1$ & $41.2$\\
 \textbf{Violet (w/ Gemini)} & $59.2$ & $21.5$ & $46.3$ & $83.2$ \\
\textbf{Violet (w/ Gemini) \textsuperscript{\scalebox{0.5}{\faSnowflake}}}& $\bf{60.3}$ & $\bf{24.8}$&  $\bf{47.2}$& $\bf{84.9}$\\
    
  \bottomrule
\end{tabular}

\caption{Results on the translated MSCoco test set. VisualGPT is trained by us on the MSCOCO dataset, and the outputs were translated using NLLB \cite{costa2022no}. (w/o Gemini) means using a normal text decoder with meshed cross-attention in each layer. \textsuperscript{\scalebox{0.5}{\faSnowflake}} indicates freezing the first part of the text decoder.}   
\label{coco}
\end{table*}

\section{Experiments}

We analyze the performance of three variations of our architecture: (i) using the normal decoder with cross-attention in each layer, (ii) using Gemini decoder without freezing the text part, and (iii) using Gemini decoder while freezing the text part. As a baseline, we train a VisualGPT model~\cite{chen2022visualgpt} on the English MSCOCO training set then translate output into Arabic using NLLB. Our trained VisualGPT achieves a $117.8$ CIDEr score on the English MSCOCO validation set. We conduct our experiments on three  datasets, as follows:

\noindent \textbf{(i) Our translated MSCOCO:} Following the Karpathy split, our translated and filtered MSCOCO contains $543,817$ samples for training (Train), $22,845$ samples for validation (Dev), and $22,912$ samples for testing (Test). We refer to this dataset as MSCOCO. 

\noindent \textbf{(ii) Translated Flickr8K:} Similar to the original Flickr8k, the translated dataset introduced in~\citet{eljundi2020resources} consists of $6,000$ images for Train, $1,000$ images for Dev, and $1,000$ for Test. Each Image has three captions, all translated using Google translate API. We refer to this dataset simply as Flickr8K.

\noindent \textbf{(iii) AraCOCO:} As described in Section~\ref{AraCoco}, AraCOCO consists of $500$ images from Karpathy test split. Each image has five captions, all obtained from human annotators. 


\subsection{Implementation Details}
We use JASMINE base ($300$m) as our text decoder. While for the detection network, following previous works~\cite{li2020oscar,cornia2020meshed,chen2022visualgpt}, we employ bottom-up attention network~\cite{anderson2018bottom} based on Resnet-101 backbone~\cite{he2016deep} with $2,048$ output features. We also limit the maximum number of detections per image to $50$ bounding boxes. The three-layer transformer encoder contains $12$ attention heads per layer with $768$ embeddings dimension.

As we utilize the JASMINE decoder~\cite{nagoudi2022jasmine}, we adopt its byte-pair encoding (BPE) vocabulary where frequent character pairs are merged to form subwords. This vocabulary encompasses $63,999$ tokens. For data preprocessing, we employ a custom normalizer that removes punctuation and repeated characters.

For the optimization part, in all experiments, we use AdamW \citet{adamw} with a learning rate of $1e^{-4}$, and empirically set $\tau$ to $0.3$. 
The model is trained using a batch size of $60$ for $20$ epochs while employing early stopping with a patience of $5$ on the validation loss. For Flickr8k, we use our MSCOCO-pretrained model and only finetune it for one epoch on Flickr8k's training data.
We employ a cross-entropy loss and train the model in an auto-regressive manner, where the decoder predicts the next token given the visual features and the previously generated textual tokens.
\subsection{Results and Discussion}




We evaluate the performance of our models against previous methods on the popular evaluation metrics BLEU~\cite{papineni2002bleu}, ROUGE~\cite{lin2004rouge}, and CIDEr~\cite{vedantam2015cider}. The results of our models on our MSCOCO dataset are displayed in Table~\ref{coco}. Our Gemini decoder with six frozen layers (last row in Table~\ref{coco}) achieves better performance while having fewer computations than the unfrozen counterpart. Furthermore, it achieves around three points higher CIDEr score compared to the translated VisualGPT outputs (first row in Table~\ref{coco}). The poor performance observed using the full decoder with cross-attention layers (second row in Table~\ref{coco}), compared to other variants may be due to sensitivity of the decoder parameters which end up being changed significantly with full cross-attention across all its layers. 
\begin{table}[htp]
\centering
 \renewcommand{\arraystretch}{1.3}
\resizebox{\columnwidth}{!}{%
\begin{tabular}{p{2.0cm}p{1.4cm}p{1.4cm}p{1.4cm}p{1.4cm}}
\toprule
\textbf{Model}&\textbf{BLEU-1}&\textbf{BLEU-4}  & \textbf{Rouge} &  \textbf{CIDEr} \\
\midrule

 \textbf{\scriptsize\newcite{elbedwehy2023improved}} & $\bf{58.7}$ & $\bf{16.5}$ & $\underline{38.0}$& $\underline{46.9}$ 
 \\
 \textbf{\scriptsize\citet{emami2022arabic}}  & $39.0$ & $09.0$ & $33.4$ & $42.3$ \\
   
    \midrule
    
   \rowcolor{violet!6}\small\textbf{ Violet} &  $\underline{44.2}$ & $\underline{13.0}$& $\bf{38.4}$&  $\bf{60.1}$ \\
    
  \bottomrule
\end{tabular}}
\caption{Results on Flickr8k test set from \cite{eljundi2020resources}. The results are taken from the respective papers.}
\label{flickr}
\end{table}

To compare our Arabic captioning model with previously published Arabic models, we evaluate our model on the Flickr8k test set from~\cite{eljundi2020resources}.
As shown in Table~\ref{flickr}, our model achieves $2$ points better score on the ROUGE metric, while having a substantial improvement over previous published results in the CIDEr metric. Our model scores $13$ points higher than the best model of the two previous models. On the other hand, our model falls behind in BLEU score against~\citet{elbedwehy2023improved}. It is worth noting, however, that we are only comparing to published results of~\citet{elbedwehy2023improved} since their model is not available (i.e., not released). They have also used the validation set of Flickr8k in their training, and applied self-critical~\cite{rennie2017self} with no mention of the target data, thus giving their model an advantage over our own model. Regardless, for image captioning, it is known that the CIDEr (where our model excels) is a more relevant evaluation metric than BLEU.
\begin{table}[htp]
\centering
 \renewcommand{\arraystretch}{1.3}
\resizebox{\columnwidth}{!}{%
\begin{tabular}{p{1.75cm}p{1.7cm}p{1.7cm}p{1.0cm}p{1.0cm}}
\toprule
Model &\bf{BLEU-1} & \bf{BLEU-4} & \bf{Rouge}   &  \bf{CIDEr} \\
\midrule

\textbf{VisualGPT} & $52.7$ & $17.6$ & $40.2$ & $58.5$ \\

    \textbf{Violet} & $\bf{54.5}$ & $\bf{19.0}$& $\bf{41.8}$&  $\bf{61.2}$ \\
    
  \bottomrule
\end{tabular}}
\caption{Results of our model against translated outputs of VisaulGPT on AraCOCO.}
\label{aracoco}
\end{table}
Finally, we score our model on our manually annotated dataset, AraCOCO. As shown in Table~\ref{aracoco}, our model again exhibits sizeable gains compared to our baseline model (i.e., the translated output of VisualGPT). This means that we cannot expect a satisfactory performance by simply taking output from a VLM trained on English data and translating it into Arabic, further corroborating our previous findings and motivating future work on developing VLM models that natively tailored to Arabic language.





\section{Conclusion}
In this paper, we introduced \textit{Violet}, an Arabic image captioning model leveraging the pretrained text decoder JASMINE. Our results demonstrated the efficacy of our Gemini decoder in enhancing performance while simultaneously reducing the number of model parameters and computations. We also presented a new method that is effective for acquiring Arabic captioning data from available English data. In addition, we manually annotated a new dataset for evaluating Arabic image captioning models. Our model outperforms all of our baselines and promises to enable benchmarking in this area. We will release our model and datasets to advance Arabic vison-language research.

\section{Limitations}
Similar to other image detection-based captioning models, the dependence on an external network to provide the visual features introduces an additional layer of complexity to the model. Since the model is not trained end to end, during inference, the visual features must first be obtained from the detection network before passing it to the vision encoder. Another limitation arises from the constraints of the training data. Since MSCOCO focuses solely on $80$ class objects, the model's applicability in real-world scenarios is restricted. In our future work, we aim to address both of these limitations to enhance Arabic models' efficiency and broaden their practical usage.

\section{Ethics Statement and Broad Impact}
\paragraph{Bridging the Gap in Multilingual Image Captioning.}
Image captioning serves as a crucial bridge between vision and language, with its applications touching numerous domains such as accessibility, education, and search engines. For a long time, the privilege of these advancements has been constrained to a handful of languages, primarily due to the lack of necessary datasets and dedicated research in other languages. Arabic, with its vast speakers and rich history, has unfortunately been left behind in this domain. Our work with \textit{Violet} seeks to rectify this disparity, providing a robust foundation for Arabic image captioning. By releasing \textit{Violet} and the datasets, we aim to invigorate research in this direction, promoting inclusivity and equal opportunity in NLP and computer vision advancements across languages.

\paragraph{Automated Data Acquisition and Transparency.}
To overcome the challenge of limited labeled data for Arabic image captioning, we employed a novel method for data acquisition using available English datasets. While this approach provides a solution, it also warrants a discussion on the accuracy, bias, and quality of the automatically acquired data. We emphasize that while our method provides a foundational dataset, manual annotations and human evaluations remain paramount for ensuring data quality and avoiding propagation of errors.


\paragraph{Acknowledgment of Data Sources and Fair Credit.}
Similar to ensuring proper credit assignment for benchmarking tasks, we emphasize the importance of acknowledging the original data sources we leveraged, especially in the context of automated data acquisition. Users and researchers utilizing our datasets and model are encouraged to cite and acknowledge the original datasets and sources. This practice ensures that original creators receive the recognition they deserve and promotes a culture of transparency and fairness in the research community.

\section*{Acknowledgments}\label{sec:acknow}
We acknowledge support from Canada Research Chairs (CRC), the Natural Sciences and Engineering Research Council of Canada (NSERC; RGPIN-2018-04267), the Social Sciences and Humanities Research Council of Canada (SSHRC; 435-2018-0576; 895-2020-1004; 895-2021-1008), Canadian Foundation for Innovation (CFI; 37771), Digital Research Alliance of Canada,\footnote{\href{https://alliancecan.ca}{https://alliancecan.ca}} and UBC ARC-Sockeye.\footnote{\href{https://arc.ubc.ca/ubc-arc-sockeye}{https://arc.ubc.ca/ubc-arc-sockeye}}
We also thank Samar Magdy, Ahmed Omar, and Karima Kadaoui for their help in creating the manually captioned subset of MSCOCO, which we refer to as AraCOCO.

\bibliography{custom}

\begin{thebibliography}{45}
\expandafter\ifx\csname natexlab\endcsname\relax\def\natexlab#1{#1}\fi

\bibitem[{Afyouni et~al.(2021)Afyouni, Azhar, and Elnagar}]{afyouni2021aracap}
Imad Afyouni, Imtinan Azhar, and Ashraf Elnagar. 2021.
\newblock Aracap: A hybrid deep learning architecture for arabic image
  captioning.
\newblock \emph{Procedia Computer Science}, 189:382--389.

\bibitem[{Al-Muzaini et~al.(2018)Al-Muzaini, Al-Yahya, and
  Benhidour}]{al2018automatic}
Huda~A Al-Muzaini, Tasniem~N Al-Yahya, and Hafida Benhidour. 2018.
\newblock Automatic arabic image captioning using rnn-lstm-based language model
  and cnn.
\newblock \emph{International Journal of Advanced Computer Science and
  Applications}, 9(6).

\bibitem[{Alayrac et~al.(2022)Alayrac, Donahue, Luc, Miech, Barr, Hasson, Lenc,
  Mensch, Millican, Reynolds et~al.}]{alayrac2022flamingo}
Jean-Baptiste Alayrac, Jeff Donahue, Pauline Luc, Antoine Miech, Iain Barr,
  Yana Hasson, Karel Lenc, Arthur Mensch, Katherine Millican, Malcolm Reynolds,
  et~al. 2022.
\newblock Flamingo: a visual language model for few-shot learning.
\newblock \emph{Advances in Neural Information Processing Systems},
  35:23716--23736.

\bibitem[{Anderson et~al.(2018)Anderson, He, Buehler, Teney, Johnson, Gould,
  and Zhang}]{anderson2018bottom}
Peter Anderson, Xiaodong He, Chris Buehler, Damien Teney, Mark Johnson, Stephen
  Gould, and Lei Zhang. 2018.
\newblock Bottom-up and top-down attention for image captioning and visual
  question answering.
\newblock In \emph{Proceedings of the IEEE conference on computer vision and
  pattern recognition}, pages 6077--6086.

\bibitem[{Attai and Elnagar(2020)}]{attai2020survey}
Anfal Attai and Ashraf Elnagar. 2020.
\newblock A survey on arabic image captioning systems using deep learning
  models.
\newblock In \emph{2020 14th International Conference on Innovations in
  Information Technology (IIT)}, pages 114--119. IEEE.

\bibitem[{Awais et~al.(2023)Awais, Naseer, Khan, Anwer, Cholakkal, Shah, Yang,
  and Khan}]{awais2023foundational}
Muhammad Awais, Muzammal Naseer, Salman Khan, Rao~Muhammad Anwer, Hisham
  Cholakkal, Mubarak Shah, Ming-Hsuan Yang, and Fahad~Shahbaz Khan. 2023.
\newblock Foundational models defining a new era in vision: A survey and
  outlook.
\newblock \emph{arXiv preprint arXiv:2307.13721}.

\bibitem[{Black et~al.(2021)Black, Gao, Wang, Leahy, and
  Biderman}]{black_sid_2021_5551208}
Sid Black, Leo Gao, Phil Wang, Connor Leahy, and Stella Biderman. 2021.
\newblock \href {https://doi.org/10.5281/zenodo.5551208} {{GPT-Neo: Large scale
  autoregressive language modeling with meshtensorflow}}.

\bibitem[{Chen et~al.(2022)Chen, Guo, Yi, Li, and
  Elhoseiny}]{chen2022visualgpt}
Jun Chen, Han Guo, Kai Yi, Boyang Li, and Mohamed Elhoseiny. 2022.
\newblock Visualgpt: Data-efficient adaptation of pretrained language models
  for image captioning.
\newblock In \emph{Proceedings of the IEEE/CVF Conference on Computer Vision
  and Pattern Recognition}, pages 18030--18040.

\bibitem[{Cornia et~al.(2020)Cornia, Stefanini, Baraldi, and
  Cucchiara}]{cornia2020meshed}
Marcella Cornia, Matteo Stefanini, Lorenzo Baraldi, and Rita Cucchiara. 2020.
\newblock Meshed-memory transformer for image captioning.
\newblock In \emph{Proceedings of the IEEE/CVF conference on computer vision
  and pattern recognition}, pages 10578--10587.

\bibitem[{Costa-juss{\`a} et~al.(2022)Costa-juss{\`a}, Cross, {\c{C}}elebi,
  Elbayad, Heafield, Heffernan, Kalbassi, Lam, Licht, Maillard
  et~al.}]{costa2022no}
Marta~R Costa-juss{\`a}, James Cross, Onur {\c{C}}elebi, Maha Elbayad, Kenneth
  Heafield, Kevin Heffernan, Elahe Kalbassi, Janice Lam, Daniel Licht, Jean
  Maillard, et~al. 2022.
\newblock No language left behind: Scaling human-centered machine translation.
\newblock \emph{arXiv preprint arXiv:2207.04672}.

\bibitem[{Elbedwehy and Medhat(2023)}]{elbedwehy2023improved}
Samar Elbedwehy and T~Medhat. 2023.
\newblock Improved arabic image captioning model using feature concatenation
  with pre-trained word embedding.
\newblock \emph{Neural Computing and Applications}, pages 1--17.

\bibitem[{ElJundi et~al.(2020)ElJundi, Dhaybi, Mokadam, Hajj, and
  Asmar}]{eljundi2020resources}
Obeida ElJundi, Mohamad Dhaybi, Kotaiba Mokadam, Hazem~M Hajj, and Daniel~C
  Asmar. 2020.
\newblock Resources and end-to-end neural network models for arabic image
  captioning.
\newblock In \emph{VISIGRAPP (5: VISAPP)}, pages 233--241.

\bibitem[{Emami et~al.(2022)Emami, Nugues, Elnagar, and
  Afyouni}]{emami2022arabic}
Jonathan Emami, Pierre Nugues, Ashraf Elnagar, and Imad Afyouni. 2022.
\newblock Arabic image captioning using pre-training of deep bidirectional
  transformers.
\newblock In \emph{Proceedings of the 15th International Conference on Natural
  Language Generation}, pages 40--51.

\bibitem[{Gan et~al.(2022)Gan, Li, Li, Wang, Liu, Gao et~al.}]{gan2022vision}
Zhe Gan, Linjie Li, Chunyuan Li, Lijuan Wang, Zicheng Liu, Jianfeng Gao, et~al.
  2022.
\newblock Vision-language pre-training: Basics, recent advances, and future
  trends.
\newblock \emph{Foundations and Trends{\textregistered} in Computer Graphics
  and Vision}, 14(3--4):163--352.

\bibitem[{Google(2023)}]{googletranslateapi2023}
Google. 2023.
\newblock \href {https://cloud.google.com/translate/docs} {Google translate
  api}.
\newblock Accessed: 15/07/2023.

\bibitem[{He et~al.(2016)He, Zhang, Ren, and Sun}]{he2016deep}
Kaiming He, Xiangyu Zhang, Shaoqing Ren, and Jian Sun. 2016.
\newblock Deep residual learning for image recognition.
\newblock In \emph{Proceedings of the IEEE conference on computer vision and
  pattern recognition}, pages 770--778.

\bibitem[{Huang et~al.(2019)Huang, Wang, Chen, and Wei}]{huang2019attention}
Lun Huang, Wenmin Wang, Jie Chen, and Xiao-Yong Wei. 2019.
\newblock Attention on attention for image captioning.
\newblock In \emph{Proceedings of the IEEE/CVF international conference on
  computer vision}, pages 4634--4643.

\bibitem[{Huang et~al.(2023)Huang, Dong, Wang, Hao, Singhal, Ma, Lv, Cui,
  Mohammed, Liu et~al.}]{huang2023language}
Shaohan Huang, Li~Dong, Wenhui Wang, Yaru Hao, Saksham Singhal, Shuming Ma,
  Tengchao Lv, Lei Cui, Owais~Khan Mohammed, Qiang Liu, et~al. 2023.
\newblock Language is not all you need: Aligning perception with language
  models.
\newblock \emph{arXiv preprint arXiv:2302.14045}.

\bibitem[{Jia et~al.(2014)Jia, Shelhamer, Donahue, Karayev, Long, Girshick,
  Guadarrama, and Darrell}]{flicker8k_caffe}
Y.~Jia, E.~Shelhamer, J.~Donahue, S.~Karayev, J.~Long, R.~Girshick,
  S.~Guadarrama, and T.~Darrell. 2014.
\newblock Caffe: Convolutional architecture for fast feature embeding.
\newblock \emph{arXiv preprint}, arXiv:1408.5093.

\bibitem[{Jindal(2017)}]{jindal2017deep}
Vasu Jindal. 2017.
\newblock A deep learning approach for arabic caption generation using
  roots-words.
\newblock In \emph{Proceedings of the AAAI Conference on Artificial
  Intelligence}, volume~31.

\bibitem[{Karpathy et~al.(2014)Karpathy, Joulin, and
  Fei-Fei}]{karpathy2014deep}
Andrej Karpathy, Armand Joulin, and Li~F Fei-Fei. 2014.
\newblock Deep fragment embeddings for bidirectional image sentence mapping.
\newblock \emph{Advances in neural information processing systems}, 27.

\bibitem[{Lasheen and Barakat(2022)}]{lasheen2022arabic}
Moaz~T Lasheen and Nahla~H Barakat. 2022.
\newblock Arabic image captioning: the effect of text pre-processing on the
  attention weights and the bleu-n scores.
\newblock \emph{Int J Adv Comput Sci Appl}, 13(7):11.

\bibitem[{Li et~al.(2011)Li, Kulkarni, Berg, Berg, and Choi}]{li2011composing}
Siming Li, Girish Kulkarni, Tamara Berg, Alexander Berg, and Yejin Choi. 2011.
\newblock Composing simple image descriptions using web-scale n-grams.
\newblock In \emph{Proceedings of the fifteenth conference on computational
  natural language learning}, pages 220--228.

\bibitem[{Li et~al.(2020)Li, Yin, Li, Zhang, Hu, Zhang, Wang, Hu, Dong, Wei
  et~al.}]{li2020oscar}
Xiujun Li, Xi~Yin, Chunyuan Li, Pengchuan Zhang, Xiaowei Hu, Lei Zhang, Lijuan
  Wang, Houdong Hu, Li~Dong, Furu Wei, et~al. 2020.
\newblock Oscar: Object-semantics aligned pre-training for vision-language
  tasks.
\newblock In \emph{Computer Vision--ECCV 2020: 16th European Conference,
  Glasgow, UK, August 23--28, 2020, Proceedings, Part XXX 16}, pages 121--137.
  Springer.

\bibitem[{Lin(2004)}]{lin2004rouge}
Chin-Yew Lin. 2004.
\newblock Rouge: A package for automatic evaluation of summaries.
\newblock In \emph{Text summarization branches out}, pages 74--81.

\bibitem[{Lin et~al.(2014)Lin, Maire, Belongie, Hays, Perona, Ramanan,
  Doll{\'a}r, and Zitnick}]{lin2014microsoft}
Tsung-Yi Lin, Michael Maire, Serge Belongie, James Hays, Pietro Perona, Deva
  Ramanan, Piotr Doll{\'a}r, and C~Lawrence Zitnick. 2014.
\newblock Microsoft coco: Common objects in context.
\newblock In \emph{Computer Vision--ECCV 2014: 13th European Conference,
  Zurich, Switzerland, September 6-12, 2014, Proceedings, Part V 13}, pages
  740--755. Springer.

\bibitem[{Loshchilov and Hutter(2019)}]{adamw}
Ilya Loshchilov and Frank Hutter. 2019.
\newblock Decoupled weight decay regularization.
\newblock \emph{In ICLR}.

\bibitem[{Nagoudi et~al.(2022)Nagoudi, Abdul-Mageed, Elmadany, Inciarte, and
  Khondaker}]{nagoudi2022jasmine}
El~Moatez~Billah Nagoudi, Muhammad Abdul-Mageed, AbdelRahim Elmadany,
  Alcides~Alcoba Inciarte, and Md~Tawkat~Islam Khondaker. 2022.
\newblock Jasmine: Arabic gpt models for few-shot learning.
\newblock \emph{arXiv preprint arXiv:2212.10755}.

\bibitem[{Ouyang et~al.(2022)Ouyang, Wu, Jiang, Almeida, Wainwright, Mishkin,
  Zhang, Agarwal, Slama, Ray et~al.}]{ouyang2022training}
Long Ouyang, Jeffrey Wu, Xu~Jiang, Diogo Almeida, Carroll Wainwright, Pamela
  Mishkin, Chong Zhang, Sandhini Agarwal, Katarina Slama, Alex Ray, et~al.
  2022.
\newblock Training language models to follow instructions with human feedback.
\newblock \emph{Advances in Neural Information Processing Systems},
  35:27730--27744.

\bibitem[{Papineni et~al.(2002)Papineni, Roukos, Ward, and
  Zhu}]{papineni2002bleu}
Kishore Papineni, Salim Roukos, Todd Ward, and Wei-Jing Zhu. 2002.
\newblock Bleu: a method for automatic evaluation of machine translation.
\newblock In \emph{Proceedings of the 40th annual meeting of the Association
  for Computational Linguistics}, pages 311--318.

\bibitem[{Plummer et~al.(2015)Plummer, Wang, Cervantes, Caicedo, Hockenmaier,
  and Lazebnik}]{plummer2015flickr30k}
Bryan~A Plummer, Liwei Wang, Chris~M Cervantes, Juan~C Caicedo, Julia
  Hockenmaier, and Svetlana Lazebnik. 2015.
\newblock Flickr30k entities: Collecting region-to-phrase correspondences for
  richer image-to-sentence models.
\newblock In \emph{Proceedings of the IEEE international conference on computer
  vision}, pages 2641--2649.

\bibitem[{Radford et~al.(2021)Radford, Kim, Hallacy, Ramesh, Goh, Agarwal,
  Sastry, Askell, Mishkin, Clark et~al.}]{radford2021learning}
Alec Radford, Jong~Wook Kim, Chris Hallacy, Aditya Ramesh, Gabriel Goh,
  Sandhini Agarwal, Girish Sastry, Amanda Askell, Pamela Mishkin, Jack Clark,
  et~al. 2021.
\newblock Learning transferable visual models from natural language
  supervision.
\newblock In \emph{International conference on machine learning}, pages
  8748--8763. PMLR.

\bibitem[{Radford et~al.(2019)Radford, Wu, Child, Luan, Amodei, Sutskever
  et~al.}]{radford2019language}
Alec Radford, Jeffrey Wu, Rewon Child, David Luan, Dario Amodei, Ilya
  Sutskever, et~al. 2019.
\newblock Language models are unsupervised multitask learners.
\newblock \emph{OpenAI blog}, 1(8):9.

\bibitem[{Reimers and Gurevych(2019)}]{reimers2019sentence}
Nils Reimers and Iryna Gurevych. 2019.
\newblock Sentence-bert: Sentence embeddings using siamese bert-networks.
\newblock \emph{arXiv preprint arXiv:1908.10084}.

\bibitem[{Ren et~al.(2015)Ren, He, Girshick, and Sun}]{ren2015faster}
Shaoqing Ren, Kaiming He, Ross Girshick, and Jian Sun. 2015.
\newblock Faster r-cnn: Towards real-time object detection with region proposal
  networks.
\newblock \emph{Advances in neural information processing systems}, 28.

\bibitem[{Rennie et~al.(2017)Rennie, Marcheret, Mroueh, Ross, and
  Goel}]{rennie2017self}
Steven~J Rennie, Etienne Marcheret, Youssef Mroueh, Jerret Ross, and Vaibhava
  Goel. 2017.
\newblock Self-critical sequence training for image captioning.
\newblock In \emph{Proceedings of the IEEE conference on computer vision and
  pattern recognition}, pages 7008--7024.

\bibitem[{Sabri(2021)}]{sabri2021arabic}
Sabri~Monaf Sabri. 2021.
\newblock \emph{Arabic image captioning using deep learning with attention}.
\newblock Ph.D. thesis, University of Georgia.

\bibitem[{Stefanini et~al.(2022)Stefanini, Cornia, Baraldi, Cascianelli,
  Fiameni, and Cucchiara}]{stefanini2022show}
Matteo Stefanini, Marcella Cornia, Lorenzo Baraldi, Silvia Cascianelli,
  Giuseppe Fiameni, and Rita Cucchiara. 2022.
\newblock From show to tell: A survey on deep learning-based image captioning.
\newblock \emph{IEEE transactions on pattern analysis and machine
  intelligence}, 45(1):539--559.

\bibitem[{Vedantam et~al.(2015)Vedantam, Lawrence~Zitnick, and
  Parikh}]{vedantam2015cider}
Ramakrishna Vedantam, C~Lawrence~Zitnick, and Devi Parikh. 2015.
\newblock Cider: Consensus-based image description evaluation.
\newblock In \emph{Proceedings of the IEEE conference on computer vision and
  pattern recognition}, pages 4566--4575.

\bibitem[{Vinyals et~al.(2015)Vinyals, Toshev, Bengio, and
  Erhan}]{vinyals2015show}
Oriol Vinyals, Alexander Toshev, Samy Bengio, and Dumitru Erhan. 2015.
\newblock Show and tell: A neural image caption generator.
\newblock In \emph{Proceedings of the IEEE conference on computer vision and
  pattern recognition}, pages 3156--3164.

\bibitem[{Wang et~al.(2022)Wang, Yang, Hu, Li, Lin, Gan, Liu, Liu, and
  Wang}]{wang2022git}
Jianfeng Wang, Zhengyuan Yang, Xiaowei Hu, Linjie Li, Kevin Lin, Zhe Gan,
  Zicheng Liu, Ce~Liu, and Lijuan Wang. 2022.
\newblock Git: A generative image-to-text transformer for vision and language.
\newblock \emph{arXiv preprint arXiv:2205.14100}.

\bibitem[{Xu et~al.(2015)Xu, Ba, Kiros, Cho, Courville, Salakhudinov, Zemel,
  and Bengio}]{xu2015show}
Kelvin Xu, Jimmy Ba, Ryan Kiros, Kyunghyun Cho, Aaron Courville, Ruslan
  Salakhudinov, Rich Zemel, and Yoshua Bengio. 2015.
\newblock Show, attend and tell: Neural image caption generation with visual
  attention.
\newblock In \emph{International conference on machine learning}, pages
  2048--2057. PMLR.

\bibitem[{Yang et~al.(2011)Yang, Teo, Daum{\'e}~III, and
  Aloimonos}]{yang2011corpus}
Yezhou Yang, Ching Teo, Hal Daum{\'e}~III, and Yiannis Aloimonos. 2011.
\newblock Corpus-guided sentence generation of natural images.
\newblock In \emph{Proceedings of the 2011 conference on empirical methods in
  natural language processing}, pages 444--454.

\bibitem[{You et~al.(2016)You, Jin, Wang, Fang, and Luo}]{you2016image}
Quanzeng You, Hailin Jin, Zhaowen Wang, Chen Fang, and Jiebo Luo. 2016.
\newblock Image captioning with semantic attention.
\newblock In \emph{Proceedings of the IEEE conference on computer vision and
  pattern recognition}, pages 4651--4659.

\bibitem[{Yu et~al.(2022)Yu, Wang, Vasudevan, Yeung, Seyedhosseini, and
  Wu}]{yu2022coca}
Jiahui Yu, Zirui Wang, Vijay Vasudevan, Legg Yeung, Mojtaba Seyedhosseini, and
  Yonghui Wu. 2022.
\newblock Coca: Contrastive captioners are image-text foundation models.
\newblock \emph{arXiv preprint arXiv:2205.01917}.

\end{thebibliography}
\bibliographystyle{acl_natbib}

\end{document}